\documentclass{article}

\usepackage{arxiv}

\usepackage[utf8]{inputenc} % allow utf-8 input
\usepackage[T1]{fontenc}    % use 8-bit T1 fonts
\usepackage[hidelinks]{hyperref}       % hyperlinks
\usepackage{url}            % simple URL typesetting
\usepackage{booktabs}       % professional-quality tables
\usepackage{amsmath}
\usepackage{amssymb}
\usepackage{amsfonts}       % blackboard math symbols
\usepackage{nicefrac}       % compact symbols for 1/2, etc.
\usepackage{microtype}      % microtypography
\usepackage{cleveref}       % smart cross-referencing
\usepackage{lipsum}         % Can be removed after putting your text content
\usepackage{graphicx}
\usepackage{float}
\usepackage{natbib}
\usepackage{doi}
\usepackage{tikz}
\usetikzlibrary{positioning}
\usetikzlibrary{calc}

\newcommand{\vect}[1]{\bold{{#1}}}

\title{A Distribution-Based Threshold for Determining Sentence Similarity}

\date{}

\newif\ifuniqueAffiliation
\uniqueAffiliationtrue

\ifuniqueAffiliation % Standard variant of author block
\author{Gioele Cadamuro\\
	ATS Relab\\
	Milan, Italy\\
	\texttt{gioele.cadamuro@atsrelab.com} \\
	\And Marco Gruppo \\
	ATS Relab\\
	Milan, Italy\\
	\texttt{marco.gruppo@atsrelab.com} \\
}
\else
\usepackage{authblk}

\newbox{\orcid}\sbox{\orcid}{\includegraphics[scale=0.06]{orcid.pdf}} 
\author[1]{%
	\href{https://orcid.org/0000-0000-0000-0000}{\usebox{\orcid}\hspace{1mm}David S.~Hippocampus\thanks{\texttt{hippo@cs.cranberry-lemon.edu}}}%
}
\author[1,2]{%
	\href{https://orcid.org/0000-0000-0000-0000}{\usebox{\orcid}\hspace{1mm}Elias D.~Striatum\thanks{\texttt{stariate@ee.mount-sheikh.edu}}}%
}
\affil[1]{Department of Computer Science, Cranberry-Lemon University, Pittsburgh, PA 15213}
\affil[2]{Department of Electrical Engineering, Mount-Sheikh University, Santa Narimana, Levand}
\fi

\renewcommand{\shorttitle}{A Distribution-Based Threshold for Determining Sentence Similarity}

\hypersetup{
pdftitle={A Distribution-Based Threshold for Determining Sentence Similarity},
pdfsubject={cs.CL},
pdfauthor={Gioele Cadamuro, Marco Gruppo},
pdfkeywords={natural language processing, sentence similarity, applications in finance},
}

\begin{document}
\maketitle

\begin{abstract}
    We hereby present a solution to a semantic textual similarity (STS) problem in which it is necessary to match two sentences containing, as the only distinguishing factor, highly specific information (such as names, addresses, identification codes), and from which we need to derive a definition for when they are similar and when they are not.\\
    The solution revolves around the use of a neural network, based on the siamese architecture, to create the distributions of the distances between similar and dissimilar pairs of sentences. The goal of these distributions is to find a discriminating factor, that we call "threshold", which represents a well-defined quantity that can be used to distinguish vector distances of similar pairs from vector distances of dissimilar pairs in new predictions and later analyses.
    In addition, we developed a way to score the predictions by combining attributes from both the distributions' features and the way the distance function works.\\
    Finally, we generalize the results showing that they can be transferred to a wider range of domains by applying the system discussed to a well-known and widely used benchmark dataset for STS problems.
\end{abstract}

% keywords can be removed

\section{Introduction}\label{sec:introduction}
Semantic textual similarity (STS) is the process of finding how similar are two given pieces of text. Usually, this similarity is yielded as a score, for example between 1 to 5 as in the SICK dataset \citep{marelli2014sick} or the STS benchmark \citep{cer2017semeval}.\\
In this paper we focus on a different analysis of the concept of similarity: through the use of vector representations and mathematical distances, we go into what it takes for two sentences to be considered similar.\\
In our dataset we are presented with a binary problem, a pair of sentences can be either similar or not similar, there is a high density of personal information and identification codes, and the sentences don't always have a clear meaning but mostly represent stacked information. The result is a very loosely defined concept of similarity.\\
We present a solution that studies the similarity through the use of the distributions of the two classes and derives a discriminating factor that determines what can be considered similar and what is not.\\
To create the distributions we rely on a siamese neural network. This type of architecture is composed of two identical input paths with tied weights\,---\,which means that we create only one instance of each layer.\\
The inputs are sequences of any length, which are then mapped to $512$-dimensional vectors belonging to the vector space $\mathbb{R}^{512}$.\\
The process of mapping from text to vector representations is handled by a pre-trained multilingual Universal Sentence Encoder \citep{yang2019multilingual}. This model uses the transformer architecture \citep{NIPS2017_3f5ee243} to create embeddings for variable text sequences and was built with transfer learning in mind.\\
More specifically, the model uses the encoder component of the transformer, along with bi-directional self-attention\,---\,for computing context-aware representations of tokens, which take into account both their ordering and identity.\\
The Universal Sentence Encoder is also available with a convolutional neural network architecture \citep{kim-2014-convolutional}, which instead uses average pooling to convert the token-level embeddings into a fixed-length representation.\\
The difference between the two architectures is summarized as follows: the transformer has a higher accuracy but consumes more resources, while the CNN has a faster inference time but lower accuracy.\\
For the results in this paper, we preferred the transformer architecture because of the higher accuracy, however, the CNN architecture may come in handy if the model is deployed in a setting that requires less consumption of resources rather than a slightly better accuracy.\\
\cite{cer2018universal} demonstrated that, when it comes to transfer learning, sentence embeddings outperforms words embeddings such as word2vec \citep{NIPS2013_9aa42b31} and GloVe \citep{pennington-etal-2014-glove}.\\
As detailed by \cite{yang2019multilingual}, the training data used for the multilingual Universal Sentence Encoder consists of mined question-answer pairs (from online forums and QA websites), mined translation pairs, and the Stanford Natural Language Inference (SNLI) corpus \citep{bowman-etal-2015-large}, which was also translated from English to the other $15$ languages that the model covers. We refer back to the cited paper for more details on the model.\\
After the two sentences have been converted into a vector representation and their most important features selected, we calculate the distance between the two representations and use the resulting number to synthesize the characteristics of the similarity into a well-defined mathematical value.\\
We restricted our research to Euclidean, Manhattan, and Minkowski distances, but others may work as well.\\
The reason why we use a distance function is that by mapping our data into a vector space the closer two points are the closer their representation, and thus the more similar we can consider the original information.\\
The paper goes on to describe a way to score the predictions of the model and therefore associate an accuracy for each output. This method works by taking into account the optimal value for each class and how the values are distributed.\\
Finally, we apply everything described to a new set of data, more specifically we fit the SICK benchmark \citep{marelli2014sick} to the boundaries of our problem, derive a threshold value, and analyze the distributions. We, therefore, show that the method described can be applied and tested in wider domains, with less specific characteristics than the one that originally inspired the work.

\section{Related Work}\label{sec:related_work}
The foundation of our work lays primarily on the concepts outlined by \cite{mueller2016siamese}.\\ The Manhattan LSTM (MaLSTM) is composed of two LSTM networks with tied weights (siamese architecture) that process the input sentences into the vector representations $h_{T_a}^{(a)}$ and $h_{T_b}^{(b)}$. These are then used as inputs to the function $g$ which computes the similarity score.
\[g\left(h_{T_a}^{(a)}, h_{T_b}^{(b)}\right) = \exp\left(-||h_{T_a}^{(a)}-h_{T_b}^{(b)}||_1\right)\]
In this case, $g$ is the Manhattan distance normalized into the range $[0,1]$ by the function $\exp(-x)$.\\

The use of a siamese architecture fits naturally the problem of computing the similarity of a pair of sentences and it is therefore extremely important for our work.\\
Our model, while sharing a similar idea, differs in structure from the MaLSTM. We chose to take advantage of transfer learning using the multilingual Universal Sentence Encoder \citep{yang2019multilingual}, and we did not use the LSTM network because it didn't provide any advantage while making the training and inference times slower. Instead, we opted for a simple dense layer to reduce the features from the pre-trained model and make the vector representations smaller before computing the similarity score.\\

Our model may appear analogous to the work done by \cite{reimers2019sentence}, however, their focus is making BERT capable of being competitive in STS tasks by using a siamese/triplet network architecture; while our main objective is to derive a mathematical definition of the concept of similarity.\\
We tried to experiment with some of the concepts from SBERT and apply them to our research, but they did not yield any significant improvements.

\section{Datasets and Data Preparation}\label{sec:data}
The data used was split into two datasets: the registry and the payments.\\
The registry contains a list of payment requests with standard information about the company or person the payment has to go to, the amount to pay, and the personal information of the person or institution that has to pay. The sentence created by the combination of this information is going to be called the target.\\
On the other hand, the payments dataset contains general wire transfer information: the amount paid, why the transfer was made, and who asked for the transfer. In this case, the sentence created by the combination of this information is going to be called the driver.\\
What we want to do is associate each row in the payments dataset with the corresponding row in the registry dataset, and by doing so we are able to identify which person made the wire transfer. This is what we call a match.\\
This matching process can be abstracted as a similarity problem: finding the correct driver-target pair means associating the correct information and detecting shared words (which also include names and identification codes).\\ 
However, even though the two datasets share similar information, since the driver is filled by humans it has no constant structure and often presents incomplete information or mistakes.\\
The data we had at our disposal for the training and the study of the distributions was just below 10000 matches, which can be very limiting when it comes to deploying the system in a real setting. As in the general trend for natural language processing tasks, we expect that the system will be better able to generalize and improve with bigger datasets.\\
However, the use of pre-trained models was sufficient to capture a broader understanding of the language, and once again proved fundamental for the success of the experiments in a domain in which the data collection and labeling appear difficult and slow.
\begin{flushleft}
Because the data we worked with didn't require much pre-processing, we applied a standard cleaning procedure and removed some specific terms often repeated, which we empirically found led to an improvement in the results. Indeed, these words are standard in the language of our domain and mostly represent boilerplate that didn't provide any significant insight into the matching process.
\end{flushleft}
\begin{flushleft}
We also had to generate synthetic wrong matches, and we did that by random sampling combinations of driver-target that we knew were wrong. Since the registry contained some drivers that weren't associated with any target, we specifically used those to create most of the wrong matches. By doing so we exposed the model to a bigger set of special words and more combinations of these.\\
The data was labeled with ones for the wrong matches, and with zeros for the right matches. This decision depends on the kind of distance being used: in the Minkowski distance the closer the value is to zero the closer the vector representations. This for example is also true for the Manhattan and Euclidean distances.
\end{flushleft}

\begin{figure}[ht]
    \centering
    \includegraphics[width=0.9\textwidth]{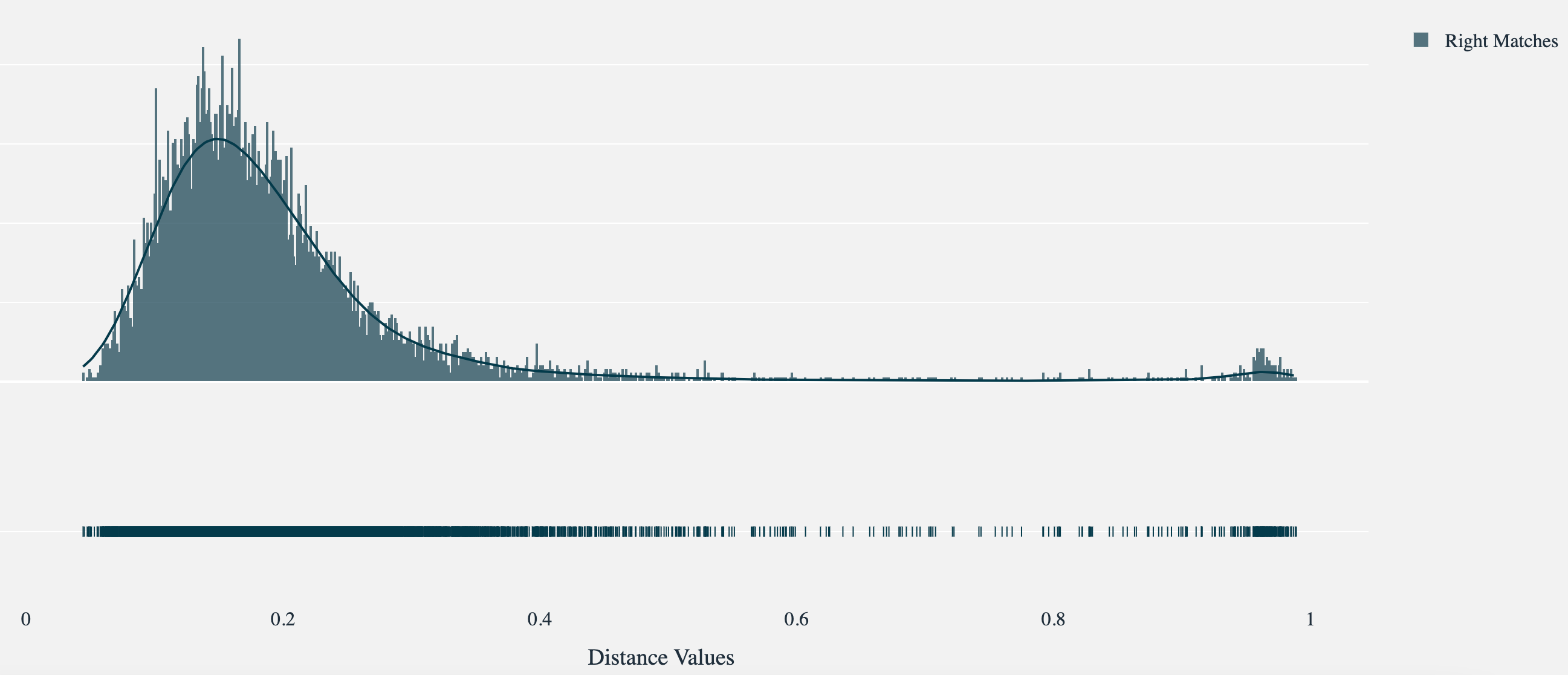}
    \caption{Distributions of right matches using the Minkowski distance.}
    \label{fig:right_distribution}
\end{figure}
\begin{figure}[ht]
    \centering
    \includegraphics[width=0.9\textwidth]{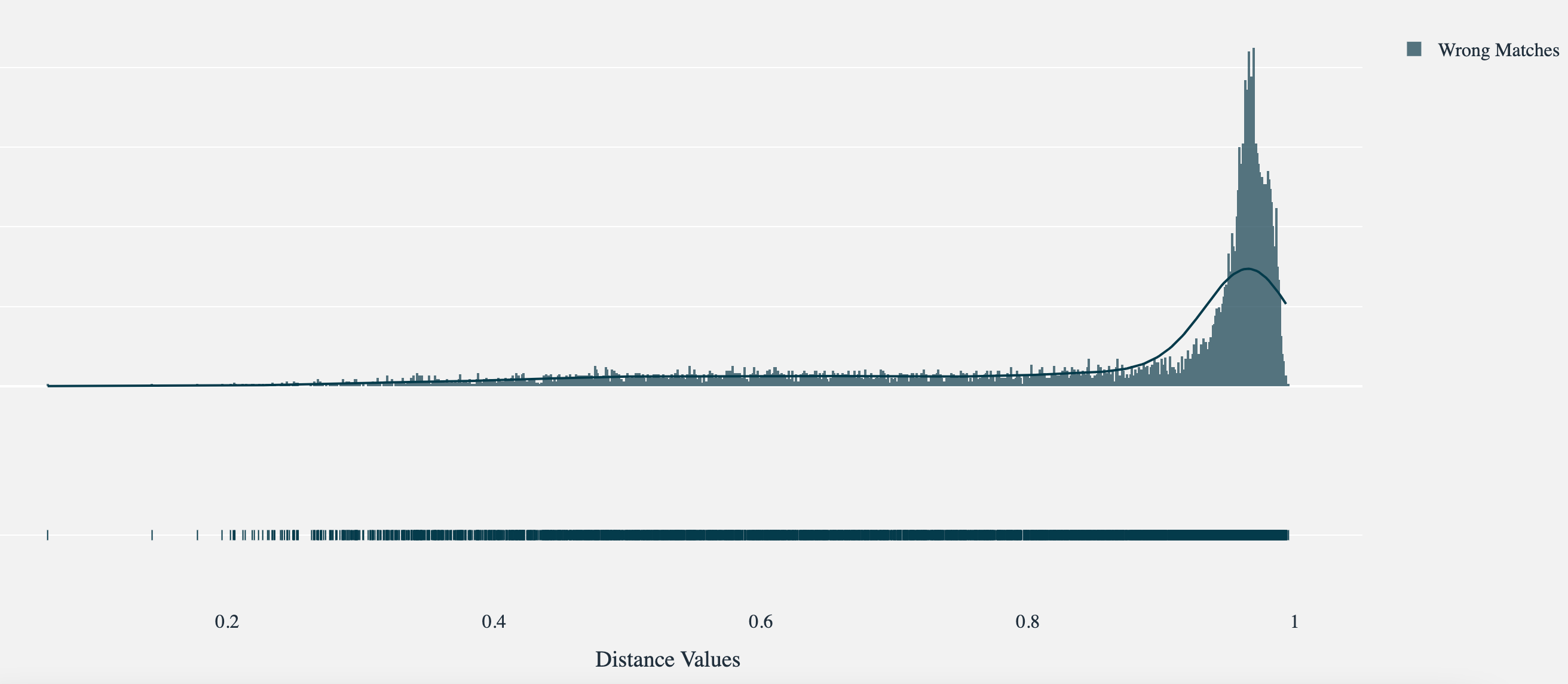}
    \caption{Distributions of wrong matches using the Minkowski distance.}
    \label{fig:wrong_distribution}
\end{figure}

\section{The Distributions and the Threshold}\label{sec:distributions_threshold}
In the early stages of our work, we were presented with a problem which is also the central theme of this paper. When can two sentences be considered similar?\\
For example, is a Minkowski distance of $0.2$ enough for two sentences to be labeled as similar? And what about a distance of $0.4$? Likewise, is a distance of $0.7$ enough for two sentences to be declared different?\\
The answer to this question was fundamental because of the domain and data we were working with. Indeed, most of the sentences we worked with didn't have any clear meaning but simply represented a bunch of information stacked together. To find a match we need to see if the information between the sentences is in some way related while navigating human mistakes, boilerplate, and words that may have no meaning for the model (e.g. small companies' names, addresses, and various codes).\\ 
In our data, there will never be a perfect match, but we will always find that the ones that we consider right cover a range of the possible distance values, and likewise the ones that we consider wrong cover another range of those values.
After training the model, we have a tool that is good enough to capture the important features that make a right match and the distinguishing features that give us a wrong match.\\
Using this model we can create the distributions of distances for our data and derive a discriminating factor for what we can consider similar and what not.\\
In figure \ref{fig:right_distribution} we can see the distribution of distance values created by the right matches from the training data and in figure \ref{fig:wrong_distribution} the distribution of the wrong matches.
\begin{figure}[ht!]
    \centering
    \includegraphics[width=0.9\textwidth]{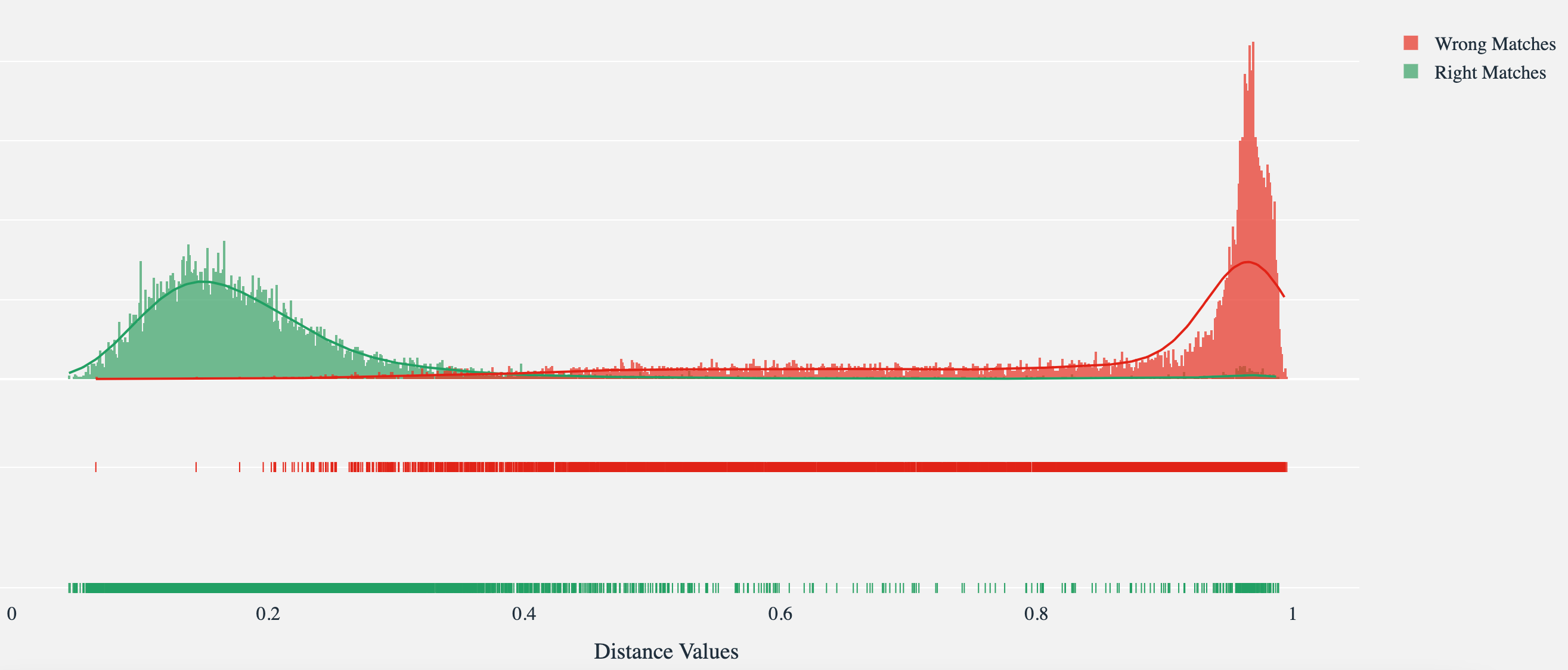}
    \caption{Distributions of right and wrong matches using the Minkowski distance.}
    \label{fig:minkowski_distributions}
\end{figure}
We can see that these distributions share some common values. Indeed it is clear from figure \ref{fig:minkowski_distributions} that the two distributions cross, which means that we have a gray area where the Minkowski distances of right and wrong matches are the same even though some are labeled as similar and others are not. This is clearly due to the nature of the data since the number of distinguishing features is low.\\
As the discriminating factor, for the distance value between what we consider wrong and what we consider right, we are going to simply take the crossing value of the two distributions, and this is what we call the threshold.\\
In this way, on the left of the threshold, we are going to have the majority of right matches, while on the right of the threshold we have the majority of wrong matches. When we have two sentences going through the model and the output is a distance whose value is less than the threshold we label that pair as similar, otherwise, we do the opposite.\\
So, we re-defined here the concept of similarity based on the position with respect to the threshold found. This allows us to generalize the model to examples of sentences where we need to find if a similarity exists, but where the concept of similarity is extremely vague.

\section{The Model}\label{sec:model}
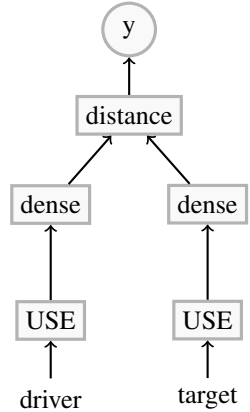
\begin{figure}[ht!]
    \centering
    \begin{tikzpicture}
        % NODES DEFINITION
        [roundnode/.style={circle, draw=gray!60, fill=gray!5, very thick, minimum size=7mm},
        squarednode/.style={rectangle, draw=gray!60, fill=gray!5, very thick, minimum size=5mm},
        textnode/.style={rectangle, draw=gray!0, fill=gray!0, very thick, minimum size=5mm}]
        
        % NODES
        % driver nodes
        \node[textnode](driver){driver};
        \node[squarednode](use_driver)[above=.5cm of driver]{USE};
        \node[squarednode](dense_driver)[above=of use_driver]{dense};
        % target nodes
        \node[textnode](target)[right=of driver]{target};
        \node[squarednode](use_target)[above=.5cm of target]{USE};
        \node[squarednode](dense_target)[above=of use_target]{dense};
        % distance and output nodes
        \coordinate (mid) at ($(driver)!0.5!(target)$);
        \node[squarednode](distance_func)[above=3.5cm of mid]{distance};
        \node[roundnode](output)[above=.5cm of distance_func]{y};
        
        % ARROWS
        % driver arrows
        \draw[->, thick] (driver.north) -- (use_driver.south);
        \draw[->, thick] (use_driver.north) -- (dense_driver.south);
        % target arrows
        \draw[->, thick] (target.north) -- (use_target.south);
        \draw[->, thick] (use_target.north) -- (dense_target.south);
        % distance and output arrows
        \draw[->, thick] (dense_driver) -- (distance_func);
        \draw[->, thick] (dense_target) -- (distance_func);
        \draw[->, thick] (distance_func) --  (output);
    \end{tikzpicture}
    \caption{Model architecture composed of the inputs (driver and target sentences), the USE layers and dense layers with shared weights, and the distance function that returns the output.}
    \label{fig:model_architecture}
\end{figure}

The model is a siamese neural network primarily based on the architecture of the Manhattan LSTM \citep{mueller2016siamese}. However, in our case, we took advantage of pre-training through the multilingual version of the Universe Sentence Encoder \citep{yang2019multilingual}, from now on abbreviated as USE, available to download from TensorFlow Hub\footnote{\url{https://tfhub.dev/google/universal-sentence-encoder-multilingual-large/3}}. Because the data was limited, the pre-training allowed us to start with a model that was already familiar with the language.\\
As already discussed, another difference with the Manhattan LSTM is that we didn't use any RNN layer\,---\,indeed it didn't improve the results and made the training slower.\\
The siamese architecture allows us to take two sentences as input and compare them directly to obtain a distance value as the output. As already seen in \cite{reimers2019sentence} this type of architecture drastically reduces computational time, and it's more efficient when finding the most similar pair.\\
After the two sentences have been processed by the USE layer, the number of features is reduced by going into a fully-connected dense layer with 50 units and the ReLU activation function.\\ Finally, the distance of the two representations is computed and returned as the output.\\
The model architecture is shown in figure \ref{fig:model_architecture}.

\subsection{Model Training}\label{subsec:model_training}
In the siamese architecture, weights are shared between the two input paths. This means that we have only one instance of the model to update.\\
We decided to keep the weights of the USE layer frozen because our training dataset was really small, and empirically it didn't seem to improve the results in any significant way.\\
After tuning the hyperparameters, the model was trained using the Adam optimizer, with a learning rate of $0.005$ and a clip normalization value of $2.0$. As discussed by \cite{pmlr-v28-pascanu13}, the gradient clipping method allows us to avoid exploding gradients and improve the training.\\
We used a batch size of $512$ and trained for $100$ epochs, but monitored the validation loss during training with early stopping.\\
Finally, we used as the loss the mean squared error, given by the following equation
\[\text{loss} = (y_\text{true}-y_\text{pred})^2\]
where $y_\text{true}$ corresponds to the ground truth label and $y_\text{pred}$ to the predicted label.\\

We obtained an $87\%$ of validation accuracy and a corresponding $0.076$ validation loss. We think these results leave some space for improvement, for example, by trying different architectures and further optimization of the hyperparameters.

\section{Distances}\label{sec:distances}
When it comes to the distance chosen, we have explored a few solutions but found the Minkowski distance to be slightly better in creating the distributions as separated as possible.\\
The choice of the distance function is important because some sentences are going to inevitably be mislabeled, therefore we need to reduce as much as possible the percentage of data that will fall into the incorrect side with respect to the threshold. To do so we have to find a distance that best fits our data by creating the smallest possible overlap between the right matches and wrong matches distributions.\\
Here we discuss three distances: the Euclidean distance, the Manhattan distance, and the Minkowski distance.\\

\textbf{Manhattan distance.} Given two points $\vect{x}=(x_1,\dots,x_n)$ and $\vect{y}=(y_1,\dots,y_n)$ the Manhattan distance corresponds to the sum of the lengths of the projections of the line segment between the vectors onto the coordinate axes
\[d(\vect{x},\vect{y}) = \sum_i^n \|x_i-y_i\|\]

\textbf{Euclidean distance.} Given two points $\vect{x}=(x_1,\dots,x_n)$ and $\vect{y}=(y_1,\dots,y_n)$ the Euclidean distance is the length of a line segment between them
\[d(\vect{x},\vect{y}) = \sqrt{\sum_i^n(x_i-y_i)^2}\]

\textbf{Minkowski distance.} Given two points $\vect{x}=(x_1,\dots,x_n)$ and $\vect{y}=(y_1,\dots,y_n)$ the Minkowski distance of order $p$ is
\[d(\vect{x},\vect{y}) = \left(\sum_i^n\|x_i-y_i\|^p\right)^{\frac{1}{p}}\]
where $p$ is an integer.\\
This metric is a generalization of both the Manhattan and the Euclidean distances: a $p$-value of $1$ gives the Manhattan distance, while a $p$-value of $2$ gives the Euclidean distance. In our implementation we used a $p$-value of $3$.\\

The output of the distance function is then taken as an input to the hyperbolic tangent function
\[\tanh\left(d(\vect{x},\vect{y})\right)\in[0,1]\]
and therefore every distance value gets squeezed between $0$ and $1$.\\

\begin{figure}[ht!]
    \centering
    \includegraphics[width=0.9\textwidth]{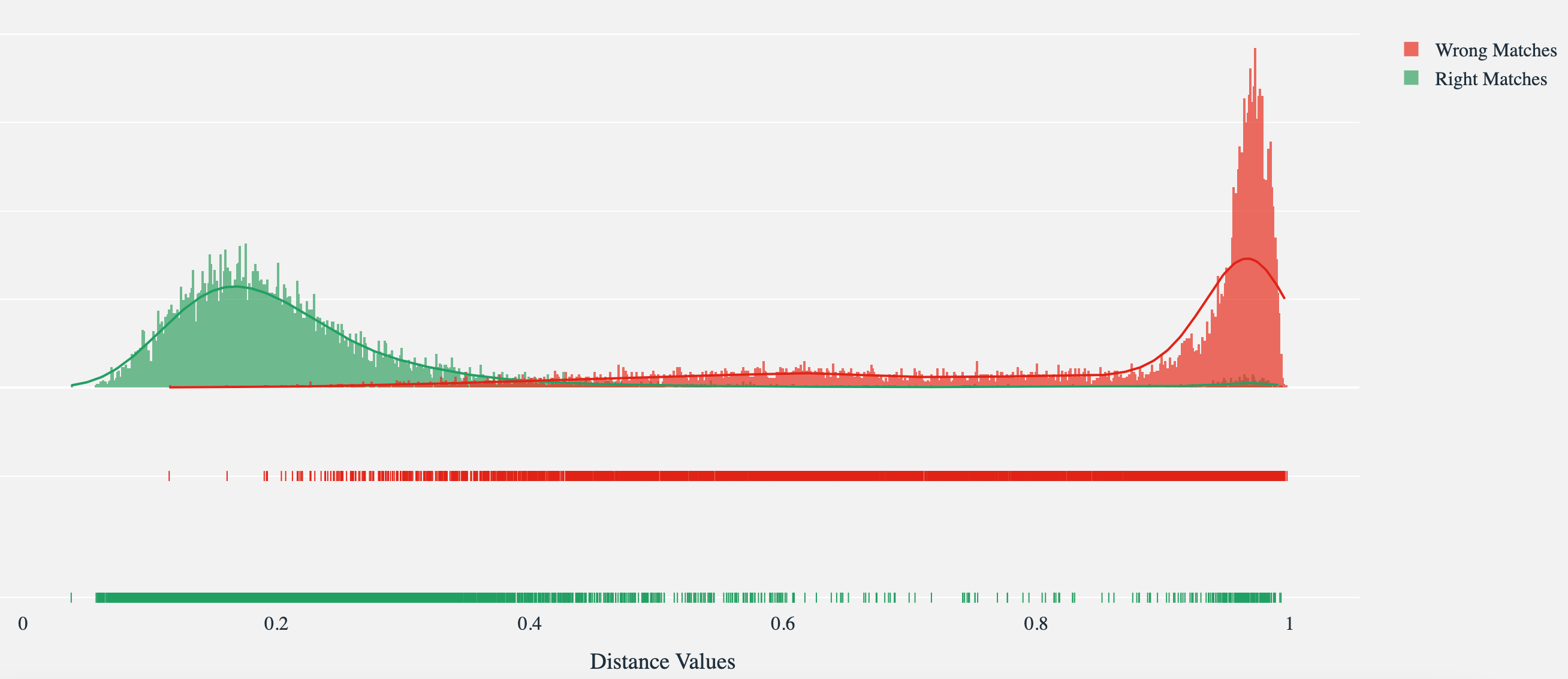}
    \caption{Distributions of right and wrong matches using the Euclidean distance.}
    \label{fig:euclidean_distributions}
\end{figure}

\begin{figure}[ht!]
    \centering
    \includegraphics[width=0.9\textwidth]{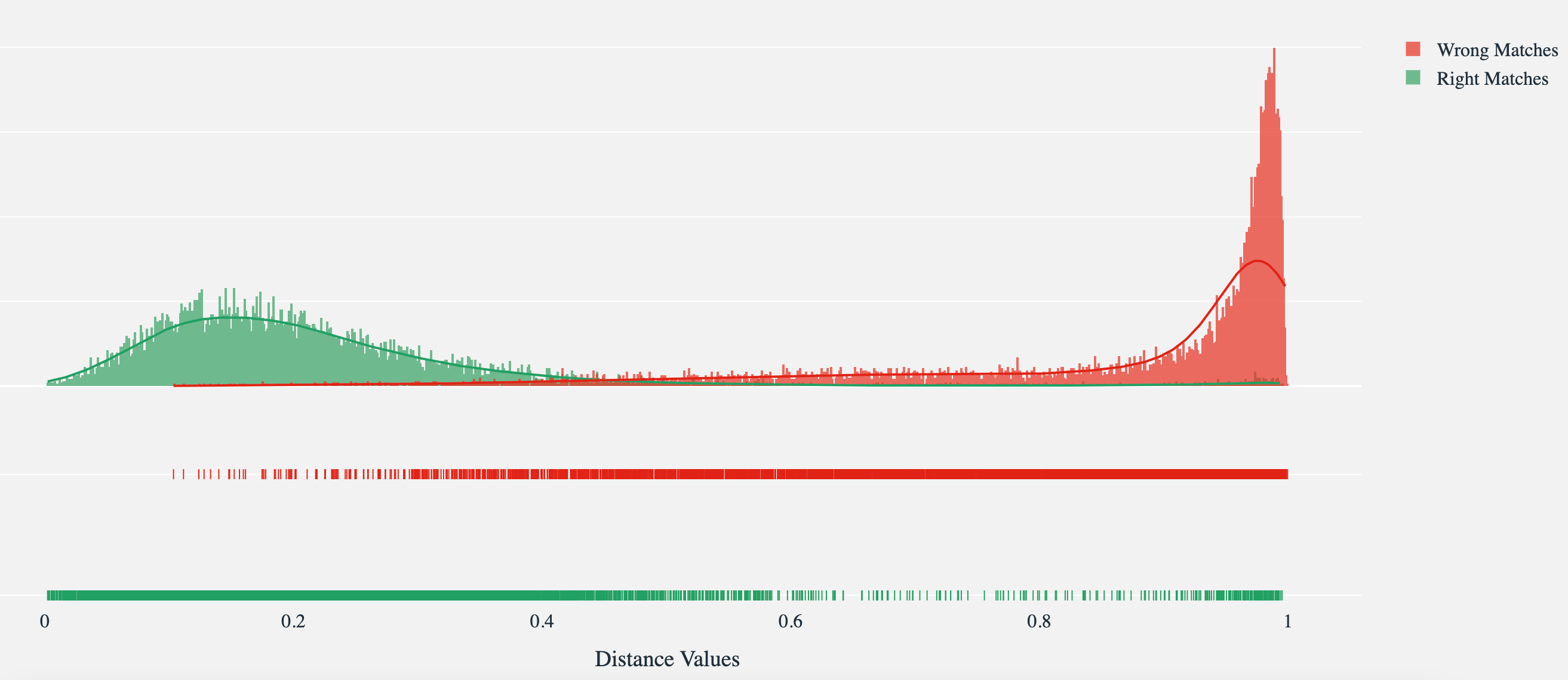}
    \caption{Distributions of right and wrong matches using the Manhattan distance.}
    \label{fig:manhattan_distributions}
\end{figure}

By looking at the figures \ref{fig:euclidean_distributions} and \ref{fig:manhattan_distributions} we can respectively see the distributions created using the Euclidean and the Manhattan distances. For the distributions using the Minkowski distance, we refer back to figure \ref{fig:minkowski_distributions}.\\
Graphically it is clear that the distributions created by the different distances don't change in a very significant way.\\
However, what we are mostly interested in is the ability to reduce mislabelled data. For the Manhattan distance, the percentage of wrong matches distances that end up on the left of the threshold and therefore in the right matches values is $4.93\%$.\\
When it comes to the Euclidean distance, this percentage is $4.56\%$. Finally, for the Minkowski distance, $4.2\%$ is the percentage of wrong matches that end up mislabeled.\\
This analysis is summarized in table \ref{tab:mislabelled_percentages}.
\begin{table}[ht!]
    \centering
    \caption{Percentage of mislabelled wrong matches for each distance function.}
    \begin{tabular}{c|c|c}
        Manhattan & Euclidean & Minkowski \\ \hline
        $4.93\%$ & $4.56\%$ & $4.2\%$
    \end{tabular}
    \label{tab:mislabelled_percentages}
\end{table}

\section{Scoring the Predictions}\label{sec:scoring_predictions}
Given a new prediction from the model and decoded its class through the threshold, how can we assign an accuracy value to it?\\
If a given prediction is labeled as right, because of the type of distance we used, we know that the closer its distance value is to zero the better the accuracy of the prediction. However, we have also noticed studying the distributions that, the peak of right matches is not at zero\,---\,due to the fact that the concept of similarity is vague.\\
Analogously, the closer the distance value is to $1.0$ the better the prediction for a wrong match, but as with right matches we also need to consider the distributions of values. Therefore, to derive an accuracy score we created what we called an "accuracy table".\\
An accuracy table takes into account both the closeness with the optimal value (for a right match that's $0.0$, for a wrong match is $1.0$) and where the majority of distance values were observed for that class, i.e. how near is it to the peak of the corresponding distribution.\\
Two accuracy tables were created, one for the right matches (i.e. anything below the threshold) and another for the wrong matches (i.e. anything above the threshold).\\
More specifically, using the distance values obtained from the data, we created a histogram with $1000$ bins containing only the values below the threshold for the right accuracy table and above the threshold for the wrong accuracy table.\\
The closeness scale is then created by associating each distance value with a bin edge number, so that the closer the distance is to the optimal value the bigger the bin edge number is. For example in the right accuracy table, because we chose $1000$ bins, the smallest possible value will have a closeness scale value of $1000$, the next smallest possible value of $999$, and so on.\\
The distribution scale, the scale built by taking into account the peak of the distribution and its properties, is created using the same histogram. In this case, we start with a value of $1.0$ and increase by $0.5$ until we get to the peak of the distribution, from which we decrease the value by $0.5$. In this way, we are taking into account the fact that there is a higher probability that a new prediction for that class will be placed close to the peak, and the farther we are from the peak the more that same probability decreases.\\

At the end of this process, we have two scales: a closeness scale and a distribution scale. To combine the information from both of them, we simply sum the corresponding values and rescale the result between $0$ and $100$ using the following equation
\[x^{(N)}_i = \frac{x_i - \min(\vect{x})}{\max(\vect{x})-\min(\vect{x})}\cdot100\]
where $\bold{x}=(x_1,\dots,x_n)$ is the array of values, $x_i$ is the $i$-th datapoint and $x^{(N)}_i$ is the corresponding $i$-th normalized datapoint.\\

We created the accuracy table, for both right and wrong matches, at the end of the training process and saved them for scoring new predictions.\\
Now, given a new prediction with a distance value and labeled as either right or wrong, to compute the accuracy of this prediction we interpolate linearly the following function
\[\text{accuracy score} = f\left(\text{distance values}\right)\]
where the independent variable is given by the distance values from the table, along with the corresponding accuracy scores for the dependent variable.\\
The implementation was provided by the SciPy library \citep{2020SciPy-NMeth}.

\section{Generalization}
The data we used to develop the system and discuss the results so far, is neither publicly available nor shareable. Therefore, even though the foundation and motivation of our work were based on the so-called registry and payments datasets previously discussed, in this section we report the results obtained by taking the concepts outlined in the paper and applying them to a well-known benchmark dataset\,---\,thus hoping to make the discussion more accessible.\\

Unfortunately, we are unaware of any dataset that presents a binary distinction between similar and not similar as in the data we used until now. So, we decided to use the SICK dataset \citep{marelli2014sick} and convert its scale of similarity into a binary problem to satisfy our system's objective.
SICK contains $10000$ pairs of sentences, and each one is classified into a $0-5$ relatedness score and whether one sentence entails the other (more information on how the data is created can be found in the original paper \cite{marelli2014sick}).\\
We consider the relatedness score and convert it into a binary label: $0$ for similar and $1$ for dissimilar.\\
Based on the way the scale was created in the first place, we decided to consider anything above a relatedness score of $3$ (excluded) to be similar and therefore have a label of $0$ and anything below $3$ (included) to have a score of $1$.\\

\begin{figure}[t]
    \centering
    \includegraphics[width=0.9\textwidth]{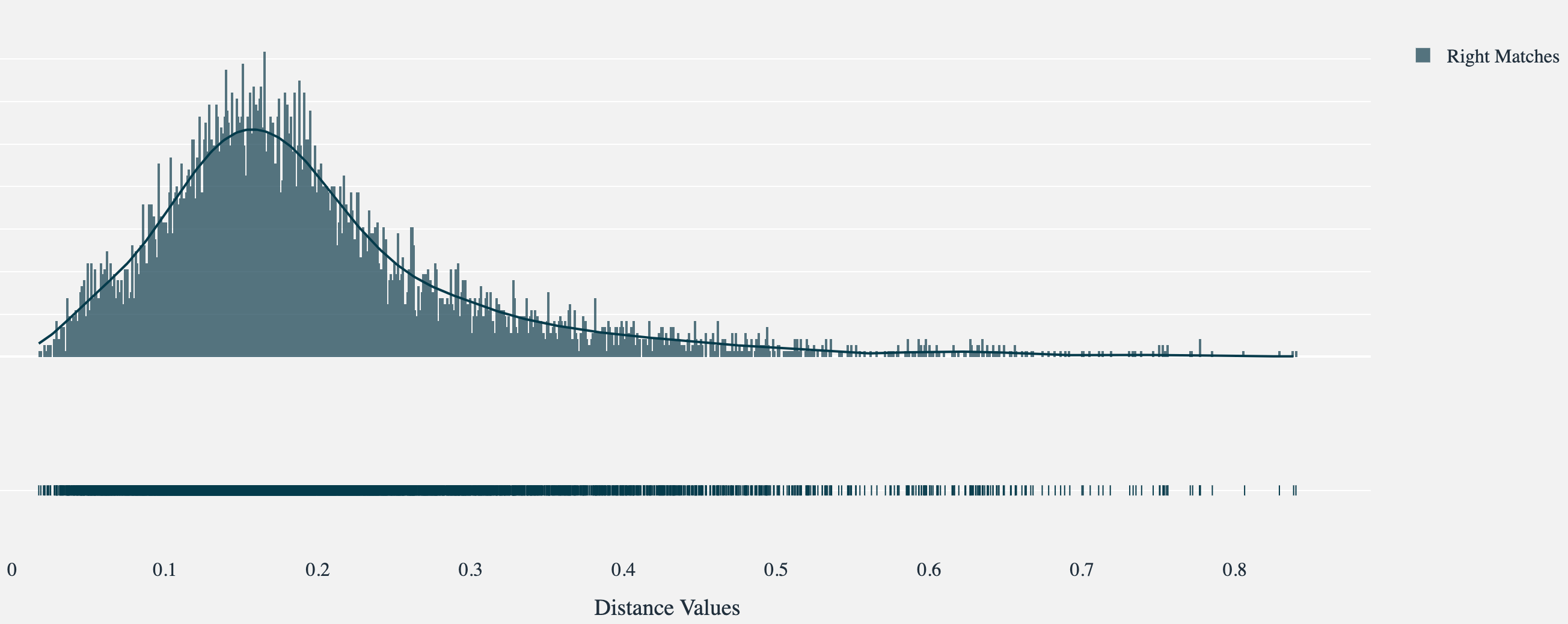}
    \caption{Distributions of right matches for the SICK dataset with the Minkowski distance.}
    \label{fig:sick_right_distribution}
\end{figure}

\begin{figure}[t]
    \centering
    \includegraphics[width=0.9\textwidth]{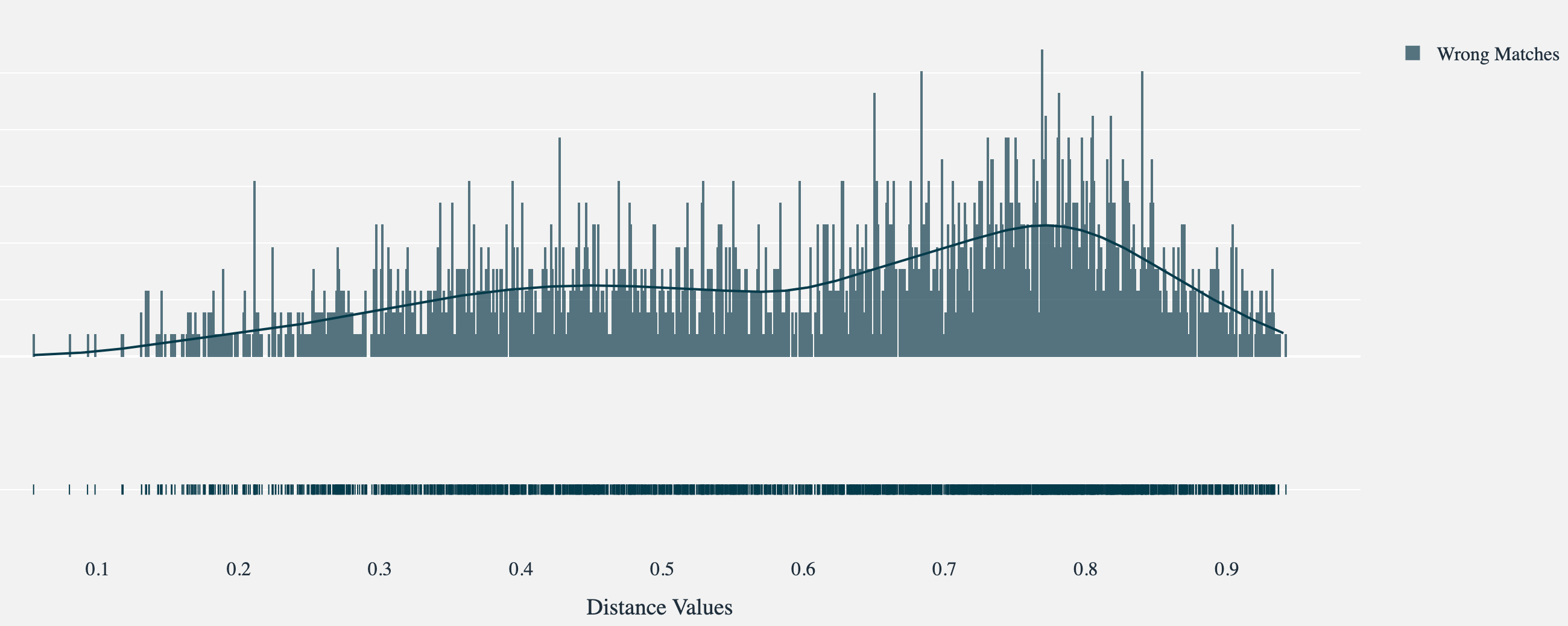}
    \caption{Distributions of wrong matches for the SICK dataset with the Minkowski distance.}
    \label{fig:sick_wrong_distribution}
\end{figure}

\begin{figure}[H]
    \centering
    \includegraphics[width=0.9\textwidth]{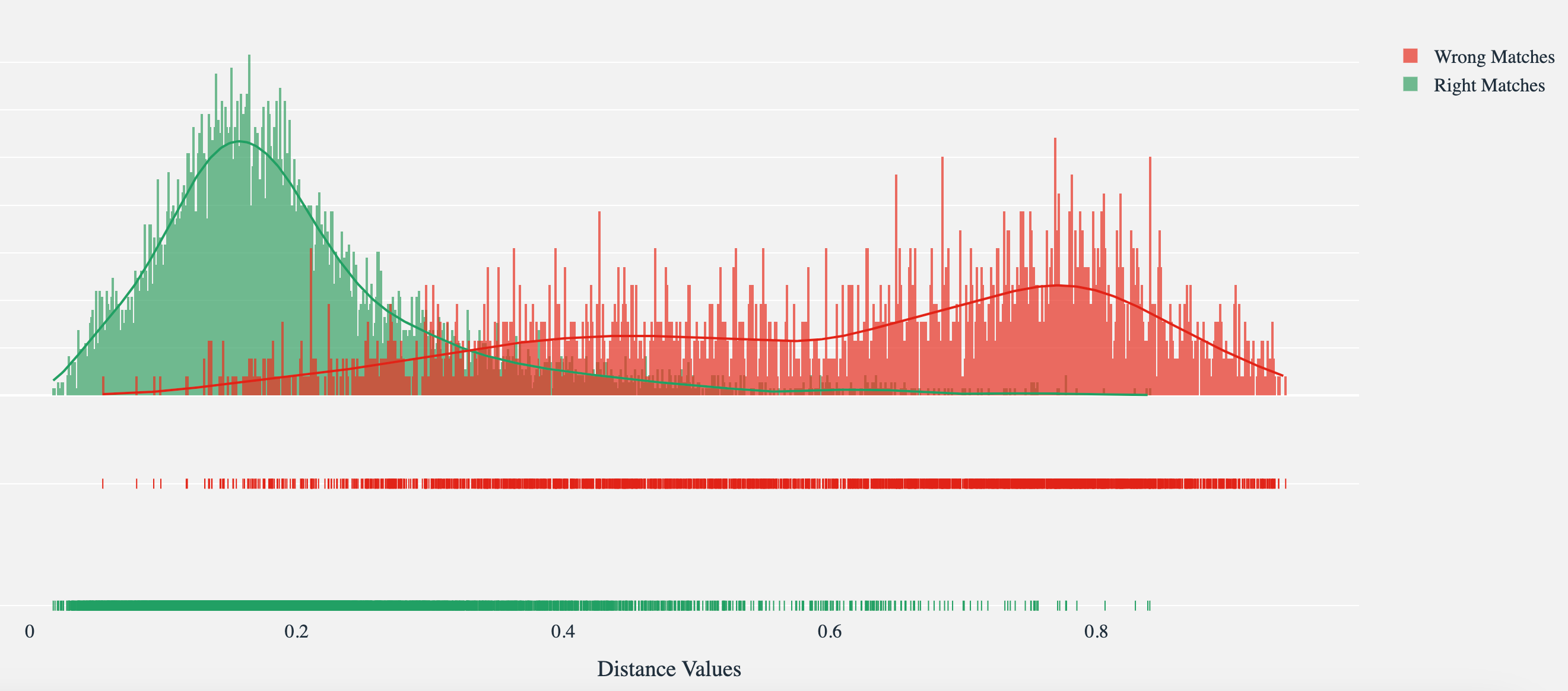}
    \caption{Distributions of right and wrong matches for the SICK dataset with the Minkowski distance.}
    \label{fig:sick_distributions}
\end{figure}

The model wasn't changed in any way, and for the training, we found the same hyperparameters discussed previously to work for this data as well.\\
We tested Euclidean, Manhattan, and Minkowski distances, and found the latter to be the best.\\
The training went on for $27$ epochs before the early stopping was triggered, and on the best epoch, we had a validation accuracy of $86.6\%$ along with a validation loss of $0.125$.\\
We report the distributions created with the SICK dataset in figures \ref{fig:sick_right_distribution} and \ref{fig:sick_wrong_distribution} separately, and together in figure \ref{fig:sick_distributions}.\\
We can see that the system is able to distinguish the two distributions, however, there is a non-insignificant portion of the wrong matches that pass through the threshold into the right matches section.\\
We calculated the percentage of mislabelled wrong matches to be $12.1\%$ which is higher than the one observed with our original data.\\
We hypothesized that the reason why this happened may be associated with the sentences that have a relatedness score in the $2-3$ range. Since we chose almost arbitrarily to put these into the wrong matches, it could be possible that some of these pairs are actually more similar than dissimilar and thus destabilize the model.\\
Indeed, we repeated the experiment by removing sentences with a relatedness score between $2-3$ and got $4\%$ of mislabelled wrong matches.\\
These results hint at the possibility of adapting and improving the system described in this paper to work on a wider range of datasets and problems.


\section{Conclusions and Comments}\label{sec:conclusions_comments}
We have discussed here a method to bring a vague concept of similarity into a well-defined quantity that can be studied and used in machine learning for solving tasks that require finding some relation in text data (and perhaps other forms of data).\\
The main interest for us was understanding what it means for two sentences to be considered similar and finding a system that could distinguish these characteristics.\\
Using a siamese neural network, and by taking advantage of pre-training, we were able to select the features that better helped us to model the concept of similarity into distributions, and finally from these distributions extract our discriminator\,---\, the threshold.\\
The final part of our work was about finding an appropriate score to associate with any new prediction, and we did that by extracting from the distributions previously created a scale of accuracy.\\

The basis for this paper was obtained using data coming from a specific domain, however, we did show that the system described can be successfully used also on more general data.\\
Even so, we want to spend the last few lines abstracting the discussion even more. The very basic idea is to have two bits of text that have some degree of similarity. As we showed here, even though the concept of similarity can be very fuzzy, it needs to be attached to something that can be extracted from the data. What we want to do next is to infer from examples of matches of similar and dissimilar pairs a discriminating factor that will allow us to classify unseen pairs of sentences into either class. As shown in this paper, distributions of the distance values can be used to accomplish this because they capture the general properties of the similarity in question.

\bibliographystyle{unsrtnat}
%\bibliography{references}  %%% Uncomment this line and comment out the ``thebibliography'' section below to use the external .bib file (using bibtex) .


\end{document}